\documentclass[11pt]{article}
\usepackage[preprint]{acl}
\usepackage{algorithm}
\usepackage{algpseudocode}
\usepackage{times}
\usepackage{latexsym}
\usepackage[T1]{fontenc}
\usepackage[utf8]{inputenc}
\usepackage{microtype}
\usepackage{inconsolata}
\usepackage{graphicx}
\usepackage{booktabs}
\usepackage{makecell}
\usepackage{placeins}
\usepackage{amsmath,amssymb}
\usepackage{xcolor}
\usepackage{colortbl}
\definecolor{tabgreen}{HTML}{D9EAD3}
\definecolor{tabred}{HTML}{F4CCCC}
\usepackage{tikz}
\usetikzlibrary{positioning}

\newcommand{\Pa}{\mathrm{Pa}}
\newcommand{\dzero}{\delta^{(0)}_{pv}}
\newcommand{\done}{\delta^{(1)}_{pv}}

\newcommand{\cref}[2]{\hyperref[#2]{#1~\ref*{#2}}}
\newcommand{\colref}[3]{\hyperref[#2]{#1~\ref*{#2}{#3}}}
\newcommand{\figref}[1]{\cref{Figure}{#1}}

\newcommand{\secref}[1]{\cref{Section}{#1}}

\newcommand{\tabref}[1]{\cref{Table}{#1}}

\definecolor{gsmheader}{HTML}{E5EFF9}
\definecolor{medheader}{HTML}{E7F2E9}
\definecolor{webheader}{HTML}{EEE8F6}
\definecolor{topologyheader}{HTML}{F0F0F0}
\definecolor{proposedwin}{HTML}{D5EEDB}
\definecolor{proposedloss}{HTML}{F8DDDD}

\title{Estimation of Propagated Risk in LLM-based Multi-Agent Systems}

\author{
  \textbf{Shreyan Ganguly}\textsuperscript{1},
  \textbf{Souradeep Chattopadhyay}\textsuperscript{1},
  \textbf{Qisai Liu}\textsuperscript{1}, \\
  \textbf{Soumik Sarkar}\textsuperscript{1} \\
  \textsuperscript{1}Iowa State University\\
  \texttt{\{shreyang, soura, supersai, soumiks\}@iastate.edu}
}

\begin{document}
\maketitle

\begin{abstract}

Large Language Model (LLM) based multi-agent systems introduce a distinctive reliability challenge in which actions taken by upstream agents can affect the correctness of downstream agents. While prior work has investigated error propagation and developed methods for estimating uncertainty inherited from upstream agents, existing approaches do not explicitly model how realized sender–receiver interactions contribute to downstream error risk under different sender correctness states. In this work, we study agent-level error-risk estimation from multi-agent execution traces. We propose Interaction-Effect Propagation (IE-Prop), an interaction-conditioned model that learns separate effects for correct and incorrect upstream agents. For each sender-receiver interaction, the model combines these conditional effects using the sender's estimated error risk, and recursively propagates the resulting risk estimates through the execution graph. We evaluate our approach across three multi-agent architectures and three benchmarks. Across all evaluated architectures, our approach raises the final-node AUROC over the established baseline by approximately 0.05 points on average. Furthermore, tracking risk across multiple agent steps contributes a distinct +0.03 AUROC gain in sequential systems. These results show that realized inter-agent interactions provide predictive information beyond receiver-local evidence, while the value of multi-step risk tracking depends on the system's information-flow structure.

\end{abstract}

\section{Introduction}
\label{sec:intro}
\begin{figure}[!t]
    \centering
    \includegraphics[width=\columnwidth]{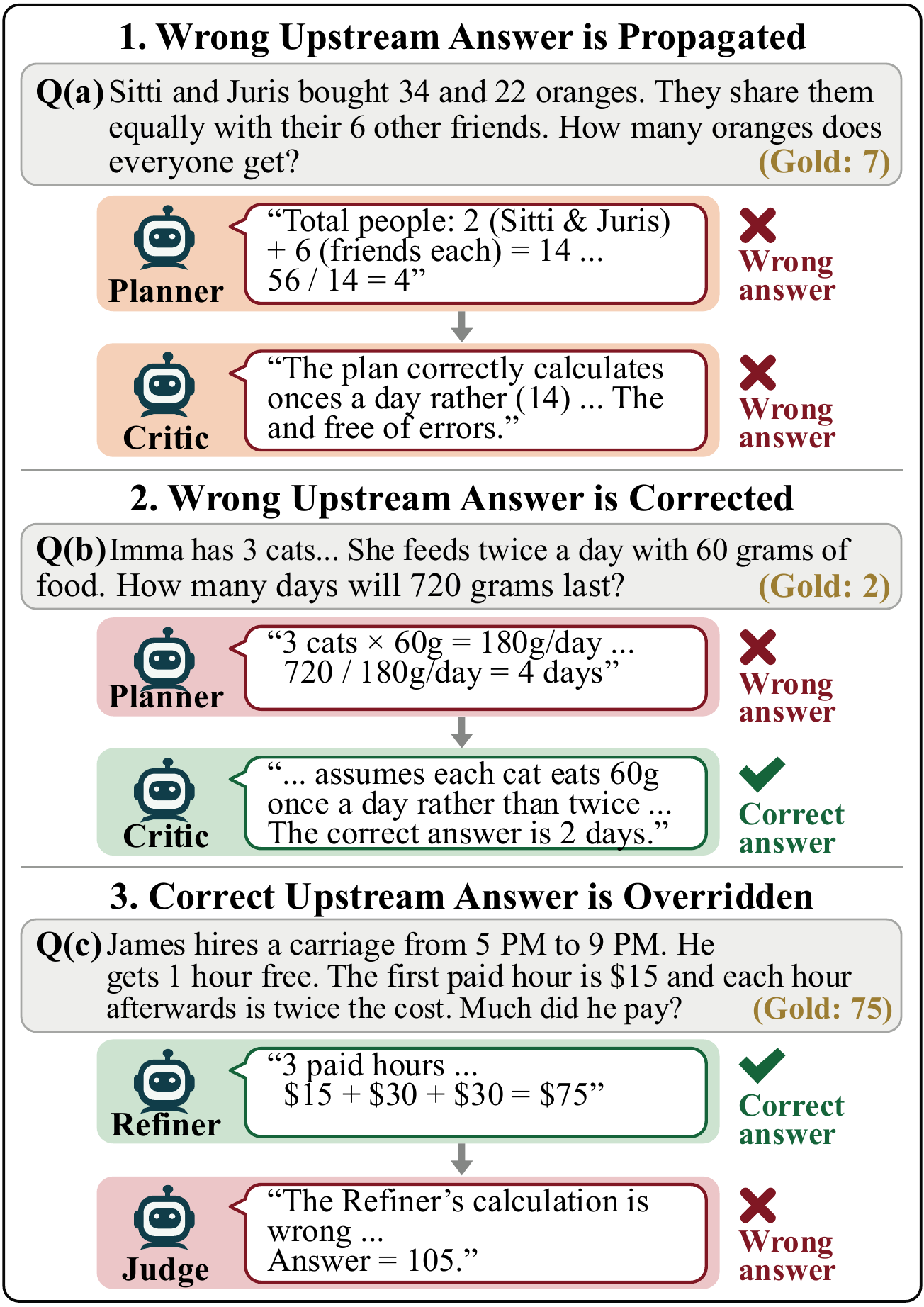}
    \caption{Error dynamics in LLM-based multi-agent systems: propagation of an upstream error (top), correction of an upstream error (middle), and erroneous override of a correct upstream answer (bottom).}
    \label{fig:figure1}
\end{figure}

LLM-based multi-agent systems decompose complex tasks across interacting agents, making intermediate outputs part of the context for subsequent generations. This interaction structure creates a reliability problem that does not arise in isolated generation: an upstream error can influence downstream predictions, but its effect depends on how the receiving agent uses the information it receives. Empirical analyses of multi-agent failures have documented error transmission through inter-agent communication \citep{cemri2025mast,zhang2025whowhen}, and controlled studies show that erroneous and corrective peer information can affect downstream behavior differently \citep{qu2026mislead,hu2026brake}. Together, these findings suggest that reliability in such systems depends not only on whether an upstream agent is likely to be wrong, but also on the realized interaction through which its output is consumed.


Existing propagation-aware approaches already account for the possibility that upstream errors are transmitted. PropUQ-MAS \citep{liu2026propuq}, for example, combines local uncertainty with inherited parent risk and uses an acceptance/critique mechanism to attenuate propagation when erroneous information is rejected or corrected. Its recurrence nevertheless treats inherited risk as a predefined contamination contribution: for fixed local uncertainty and edge acceptance, increasing parent risk cannot lower the receiver's estimated risk. This leaves open a more general question, whether the association between an observed sender-receiver interaction and receiver error should depend on the sender's correctness.

We introduce Interaction-Effect Propagation (IE-Prop), post-hoc agent-level error-risk estimation from completed multi-agent execution traces. Such risk estimates can support selective prediction and deferral and targeted re-runs \citep{elyaniv2010foundations,geifman2017selective}.
For post-hoc risk estimation, we distinguish two increasingly rich information sets available in a completed trace. First, receiver-local uncertainty signals such as entropy, verbalized uncertainty, provide a conventional estimate of the receiver's reliability. Second, a multi-agent trace exposes relational information unavailable in isolated generation: the messages received from upstream agents and the receiver's response to them. For each such interaction, we refer to the agent providing information as the sender and the agent consuming it as the receiver. Our primary empirical question is whether upstream interaction information provides predictive signal beyond the receiver text.

\begin{table}[t]
\centering
\caption{Comparison of LLM error-risk estimation methods by multi-agent system (MAS) support, risk propagation (Prop.), and underlying estimation mechanism. }
\label{tab:risk_literature_taxonomy}
\begingroup
\footnotesize
\setlength{\tabcolsep}{2.8pt}
\renewcommand{\arraystretch}{1.03}
\newcommand{\mcell}[2]{%
  \parbox[t]{.50\columnwidth}{%
    \raggedright #1\\[-1pt]{\scriptsize #2}}}
\begin{tabular}{@{}
  >{\raggedright\arraybackslash}p{.50\columnwidth}
  >{\centering\arraybackslash}p{.07\columnwidth}
  >{\centering\arraybackslash}p{.09\columnwidth}
  >{\raggedright\arraybackslash}p{.26\columnwidth}@{}}
\toprule
\textbf{Method} & \textbf{MAS} & \textbf{Prop.} & \textbf{Estimator} \\
\midrule
\mcell{P(True)}{\citep{kadavath2022know}}
  & -- & -- & Self-eval. \\
\mcell{SelfCheckGPT}{\citep{manakul2023selfcheck}}
  & -- & -- & Sample consistency \\
\mcell{Semantic Entropy}{\citep{farquhar2024detecting}}
  & -- & -- & Semantic entropy \\
\midrule
\mcell{BTProp}{\citep{hou2025btprop}}
  & -- & \checkmark & Belief tree \\
\mcell{SAUP}{\citep{zhao2025saupacl}}
  & -- & \checkmark & Situation weights \\
\mcell{UProp}{\citep{duan2025uprop}}
  & -- & \checkmark & Information theory \\
\mcell{RUPA}{\citep{ma2026rupa}}
  & -- & \checkmark & Relational graph \\
\midrule
\mcell{MATU}{\citep{chen2026matu}}
  & \checkmark & -- & Tensor factor \\
\mcell{Score Deviation}{\citep{kostka2026scoredeviation}}
  & \checkmark & -- & Conformal calibration \\
\mcell{HalluProp}{\citep{lin2026agentsspeakprehocfailure}}
  & \checkmark & \checkmark & Noisy-OR \\
\mcell{PropUQ-MAS}{\citep{liu2026propuq}}
  & \checkmark & \checkmark & Weighted Noisy-OR \\
\midrule
\textbf{Interaction-Effect Propagation (ours)}
  & \checkmark & \checkmark
  & \textbf{Signed logit mix} \\
\bottomrule
\end{tabular}
\endgroup
\end{table}

We address this question with an interaction-conditioned error-risk model. For each realized interaction, the model learns two effects on the receiver's error score, corresponding to a correct and an incorrect sender. Because sender correctness is unknown at prediction time, the two effects are combined using the sender's estimated error risk. The resulting interaction adjustment is added to receiver-local evidence, and node risks are computed recursively through the execution graph. This formulation allows the same observed interaction to have different, and potentially opposite, associations with receiver error under the two sender states, while retaining a single risk estimate that can be propagated downstream.
The contributions of our work are:

\begin{figure*}[t]
\centering
\includegraphics[width=\textwidth]{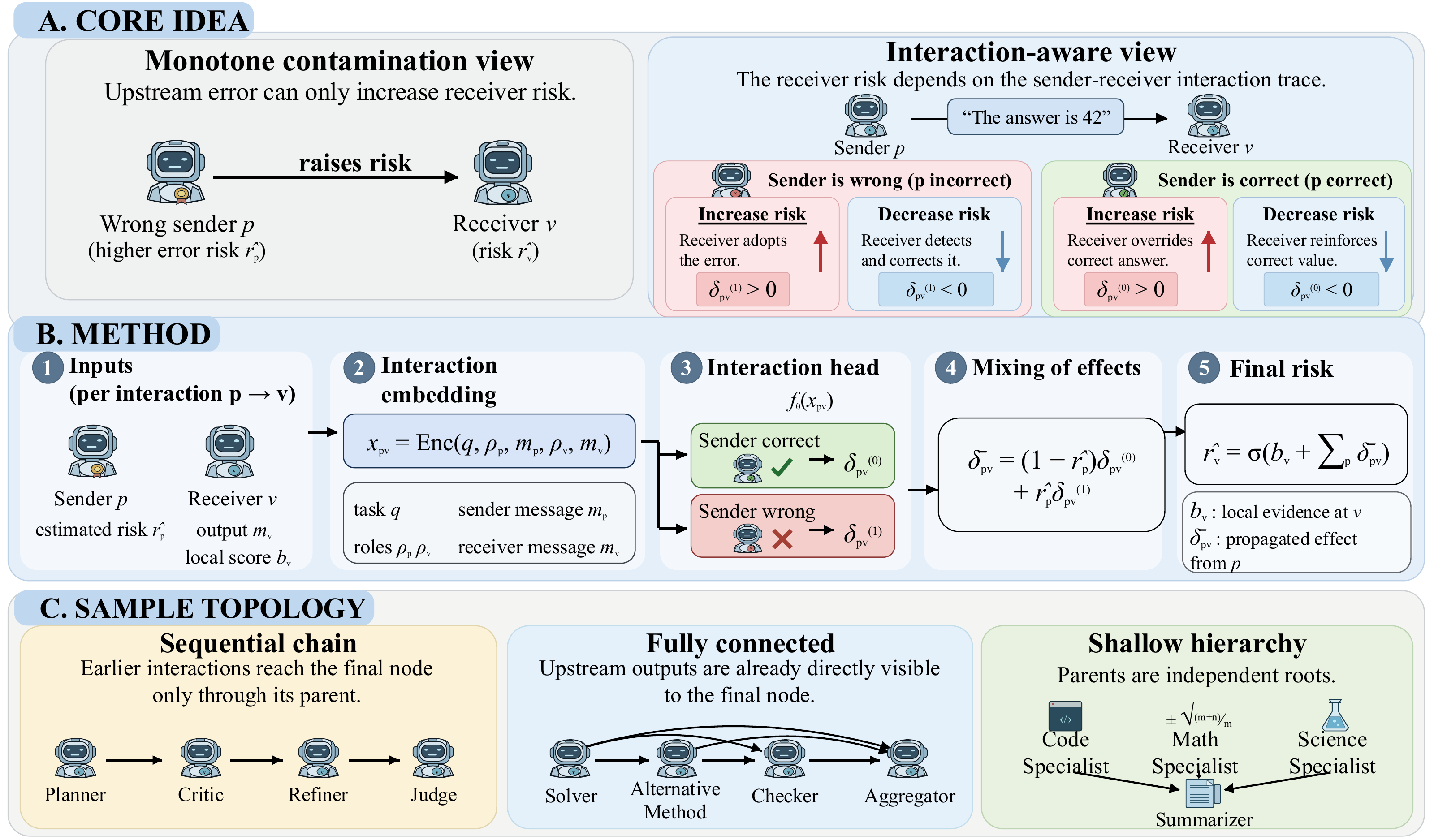}
\caption{Overview of Interaction-Effect Propagation. (A) Downstream error risk depends on realized sender–receiver interactions rather than monotone error propagation. (B) Interaction effects conditioned on sender correctness are combined using estimated sender risk and receiver-local evidence. (C) Three evaluated multi-agent topologies: sequential, fully connected, and shallow hierarchy.}
\label{fig:overview}
\end{figure*}

\textbf{(1) Interaction-conditioned, sender-state-aware error modeling} (\S\ref{sec:method}).
We introduce an agent-level error-risk model (IE-Prop) in which each realized sender-receiver interaction has separate signed effects under correct and incorrect sender states. This allows the downstream association of an interaction to depend on both what occurred between the agents and the sender's correctness.

\textbf{(2) Recursive risk propagation and learning} (\S\ref{sec:method}, \S\ref{sec:learning}).
We develop a tractable recursive procedure that combines the two interaction effects using estimated sender risk and propagates the resulting scalar risk through the execution graph. The estimator is trained end to end from node-level correctness labels, allowing a downstream node to incorporate information accumulated through earlier interactions without explicitly maintaining a joint distribution over upstream error states.

\textbf{(3) Empirical characterization of interaction and propagation effects} (\S\ref{sec:results}).
Through controlled comparisons across three multi-agent architectures, we examine whether realized upstream interactions provide predictive information beyond receiver-local evidence and distinguish the contributions of interaction modeling and recursive risk propagation. We find that interaction information generally improves downstream error-risk discrimination, although the magnitude varies across architectures and datasets.

Our results indicate downstream error risk is a property of realized inter-agent interactions, while showing that the size and the source of the benefit depend on the graph, the data and the strength of the receiver's own evidence.


\section{Related Work}
\label{sec:related}

\paragraph{LLM confidence and error-risk estimation.}
Local signals include verbalized confidence \citep{lin2022words,tian2023calibration,xiong2024express}, self-evaluation \citep{kadavath2022know}, semantic entropy \citep{kuhn2023semantic,farquhar2024detecting}, sampling consistency \citep{manakul2023selfcheck}, and trained correctness predictors \citep{mielke2022linguistic,azaria2023lying,kapoor2024taught}; benchmarks such as LM-Polygraph compare them \citep{fadeeva2023polygraph,vashurin2025benchmark}. BTProp estimates claim factuality by propagating beliefs over a tree of logically related statements \citep{hou2025btprop}. These methods estimate the uncertainty or confidence in LLM response for an isolated generation. 


\paragraph{Uncertainty and reliability in agent systems.}
SAUP propagates uncertainty across steps in a single-agent trajectory using situational weights \citep{zhao2025saupacl}, UProp adds a mutual-information term with earlier decisions \citep{duan2025uprop}, RUPA aggregates over a relation-typed trajectory graph \citep{ma2026rupa}, and MATU derives a system-level score from re-sampled multi-agent runs \citep{chen2026matu}; surveys argue that agent UQ needs formulations beyond single outputs \citep{kirchhof2025position,oh2026agentuq}.
These works study uncertainty over sequential trajectories or whole-system executions. Information preservation during inter-agent communication has also been investigated in decentralized settings \citep{rillo2026adkoagenticdecentralizedknowledge}.

\paragraph{Error propagation in multi-agent systems.}
PropUQ-MAS is the closest method to ours \citep{liu2026propuq}.
It treats an upstream error as a contamination event: each agent's local uncertainty is combined with a self-reported acceptance probability for every incoming message in a noisy-OR product, so transmission is monotone and a parent error can only increase receiver risk.
We do not assign a single transmission probability to a parent error.
Rather we learn separate effects of the realized sender-receiver interaction under correct and erroneous sender states, which allows both harmful propagation and correction. \citep{lin2026agentsspeakprehocfailure} introduce HalluProp, which estimates agent-level hallucination risk before execution using role–query alignment and topology-aware Noisy-OR propagation.
Descriptive studies measure how hallucinations survive handoffs \citep{singh2026snowball,jamshidi2026cascade}, and failure-attribution methods locate the responsible agent after a failure \citep{zhang2025whowhen,hou2026edge}.

\paragraph{Graphical-model connection.}
Our formulation is related to logistic graphical models in which node-local and relational terms combine additively in logit space \citep{neal1992connectionist,besag1974spatial}. We adopt this structure as an inductive bias, allowing each incoming interaction to contribute a separate, correctness-conditioned adjustment to the receiver's error log-odds. This provides a compositional formulation for propagating risk without explicitly maintaining a joint distribution over agents' error states (Appendix~\ref{app:graphical}).

\section{Interaction-Effect Error-Risk Estimation}
\label{sec:method}

\paragraph{Problem setup.}
We represent an executed multi-agent system as a directed
acyclic graph $G=(V,\mathcal{E})$, where $V$ is the set of
agents in the execution and
$\mathcal{E}\subseteq V\times V$ is the set of directed
edges representing realized information flow between agents.
An edge $(p,v)\in\mathcal{E}$ indicates that agent $v$
receives the visible output of agent $p$.
Each agent $v$ has a role $\rho_v$, a visible output $m_v$ and, after execution, an error indicator $E_v\in\{0,1\}$, and its error risk is $r_v=P(E_v=1)$. \(E_v\) is observed for training and evaluation traces but is unknown at inference.
Let $q$ be the task.
Each agent has a local risk score
\begin{equation}
  b_v=g_\phi(w_v),
  \label{eq:local}
\end{equation}
where $w_v$ holds agent-local risk signal and $g_\phi$ is a small network. For an interaction \(p\rightarrow v\), we refer to \(p\) as the sender (or parent) and \(v\) as the receiver, and their visible outputs are \(m_p\) and \(m_v\), respectively. Each realized interaction is embedded by a frozen text encoder
\begin{equation}
  x_{pv}=\operatorname{Enc}(q,\rho_p,m_p,\rho_v,m_v)
  \label{eq:edge}
\end{equation}
The embedding \(x_{pv}\) represents the realized exchange rather than only the parent message. It can therefore capture textual evidence of how the receiver used the upstream information - for example, whether it followed, rejected, revised, or amplified it. Because \(m_v\) is required to construct \(x_{pv}\), the estimator is necessarily post-hoc and can only score an interaction after receiver has produced its output.

\paragraph{State-conditioned interaction model.}
The association between a realized sender–receiver interaction and receiver error may differ depending on whether the sender is correct or incorrect. To capture this distinction, an edge head ($f_\theta$) maps ($x_{pv}$) to two scalars
\begin{equation}
  f_\theta(x_{pv})=\big(\dzero,\ \done\big).
\end{equation}
$\dzero$ is the effect of the realized interaction on the receiver's error log-odds if the sender is correct ($E_p=0$), and $\done$ its effect if the sender is wrong ($E_p=1$).
Positive values raise the receiver's error log-odds and negative values lower them.
The pair can therefore represent both propagation and correction: $\done>0$ describes a receiver that adopts an erroneous parent message, and $\done<0$ a receiver that detects and corrects it.

We combine receiver-local and interaction-specific evidence additively in log-odds space. The local score \(b_v\) provides the receiver's baseline error log-odds, while each incoming interaction contributes an adjustment selected by the corresponding sender state. For an error configuration \(e=\{e_p:p\in\mathrm{Pa}(v)\}\) , where \(e_p\in\{0,1\}\) indicates whether sender \(p\) is correct (\(0\)) or incorrect (\(1\)), we define
\begin{equation}
\label{eq:add}
\eta_v(e)
=
b_v+\sum_{p\in\mathrm{Pa}(v)}
\delta^{(e_p)}_{pv}.
\end{equation}
and model the receiver's conditional error probability as
\begin{equation}
\label{eq:cpd}
P(E_v=1\mid E_{\Pa(v)}=e)
=
\sigma\bigl(\eta_v(e)\bigr).
\end{equation}

This additive logistic form is a modeling choice: it provides a compositional inductive bias in which incoming interactions contribute separately to the receiver's log-odds. Appendix~\ref{app:graphical} gives the corresponding graphical-model interpretation.

\paragraph{Inference with uncertain sender states.}
Eq.~\ref{eq:cpd} specifies the receiver risk if the correctness state of every sender were known. At inference these states are unknown, so for each incoming edge we use the estimated sender error probability \(\hat r_p\) as a soft belief over the two sender states and average the corresponding interaction contributions in logit space:
\begin{equation}
\bar\delta_{pv}
=
(1-\hat r_p)\delta^{(0)}_{pv}
+
\hat r_p\delta^{(1)}_{pv}.
\label{eq:mix}
\end{equation}

The receiver's predicted risk is obtained by adding these expected interaction contributions to its local score,

\begin{equation}
\hat r_v
=
\sigma\left(
b_v+\sum_{p\in\mathrm{Pa}(v)}\bar\delta_{pv}
\right).
\label{eq:np}
\end{equation}

This averaging can be expressed in terms of expectations. Specifically, Eq.~\ref{eq:mix} computes $\bar{\delta}{pv}=\mathbb{E}_{E_p}[\delta_{pv}^{(E_p)}]$, where the expectation is taken over the sender's error state using its estimated error probability $\hat r_p$. By linearity of expectation, Eq.~\ref{eq:np} then applies the sigmoid to the expected receiver logit, yielding $\hat r_v=\sigma(\mathbb{E}_{E_{\mathrm{Pa}(v)}}[\eta_v(E_{\mathrm{Pa}(v)})])$.

Risk estimates are computed recursively in topological order over the execution graph. Roots receive $\hat r_v=\sigma(b_v)$, while each non-root node uses its parents' recursively estimated risks in Eq.~\ref{eq:mix} - ~\ref{eq:np}. So in a sequential graph a node's risk depends on interactions further upstream through its parent's risk. Appendix~\ref{app:kl} shows that this update also admits a reverse-KL barycenter interpretation.

\paragraph{Approximation to marginal inference.}
Eq.~\ref{eq:np} is a compositional inductive bias, not an exact identity: with  $Z_v=\eta_v(E_{\mathrm{Pa}(v)})$, exact marginalization of Eq.~\ref{eq:cpd} over the unknown parent states would compute $\mathbb{E}[\sigma(Z_v)]$, whereas Eq.~\ref{eq:np} computes $\sigma(\mathbb{E}[Z_v])$ from one scalar per parent, at cost $O(k)$ instead of $2^k$ for $k$ parents, and is exact when the two effects of every uncertain edge coincide.
Appendix~\ref{app:proof} derives a variance-based bound on the local discrepancy between exact marginalization and Eq.~\ref{eq:np}. Appendix~\ref{app:graphical} describes the induced probabilistic model. Either way, $\hat r_p$ enters only as the weight that mixes the two state-conditional effects, so how it is used is fixed by the model rather than learned.

\section{Training}
\label{sec:learning}

We train models $f_\theta(x_{pv})$ and $g_\phi(w_v)$, in an end-to-end setting, on execution traces generated from training questions. Each trace contains the observed interaction within agents. Predictions are computed recursively using Eqs.~\ref{eq:mix}--\ref{eq:np}, and the model parameters \(\phi\) and \(\theta\) are optimized jointly by minimizing binary cross-entropy between the predicted risk \(\hat r_v\) and the correctness label \(E_v\), summed over labeled nodes:
\begin{equation}
\label{eq:loss}
\mathcal{L}
=
\sum_{v\in V}
\operatorname{BCE}(\hat r_v,E_v).
\end{equation}
Gradients therefore propagate through the recursive risk computation, allowing both the receiver-local model \(g_\phi\) and the interaction model \(f_\theta\) to be learned jointly. Evaluation is performed on execution traces from held-out questions (\S\ref{sec:setup}). Algorithm~\ref{alg:ie} summarizes the complete procedure for recursive risk estimation, end-to-end training, and K-fold prediction.

\section{Evaluation Design}
\label{sec:ladder}

We evaluate four estimators under common training and evaluation protocol discussed in details in \S\ref{sec:setup}.

\textbf{Receiver-Only.} The local estimator $\sigma(b_v)$ (Eq.~\ref{eq:local}) predicts the receiver's error risk using three local uncertainty signals: verbalized uncertainty (\emph{verb.}), Maximum Sequence Probability (\emph{MSP}), prediction entropy.

\textbf{Receiver-Text.} Building on the Receiver-Only model, this baseline adds context about the receiver's specific task, role, and final text output, but strictly ignores any incoming messages from the sender. We use this as our main benchmark to determine whether analyzing the actual upstream conversation gives us better error predictions than just looking at the receiver in isolation.

\textbf{Free learner.} An unconstrained edge model that directly estimates the interaction adjustment $\bar{\delta}{pv}$ from the interaction representation $x_{pv}$. Unlike our proposed model, it does not decompose the adjustment into sender-state-conditioned effects or prescribe how the parent risk is used.

\begin{table*}[tp]
\centering\small
\setlength{\tabcolsep}{2.5pt}
\resizebox{\textwidth}{!}{\begin{tabular}{l||ccc|ccc|ccc||ccc|ccc|ccc}
\toprule
 & \multicolumn{9}{c||}{\cellcolor{gsmheader}GSM8K} & \multicolumn{9}{c}{\cellcolor{medheader}MedQA} \\
\cmidrule(lr){2-10}\cmidrule(lr){11-19}
 & \multicolumn{3}{c|}{Verb.} & \multicolumn{3}{c|}{MSP} & \multicolumn{3}{c||}{Entropy} & \multicolumn{3}{c|}{Verb.} & \multicolumn{3}{c|}{MSP} & \multicolumn{3}{c}{Entropy} \\
\cmidrule(lr){2-4}\cmidrule(lr){5-7}\cmidrule(lr){8-10}\cmidrule(lr){11-13}\cmidrule(lr){14-16}\cmidrule(lr){17-19}
Estimator & 0.6B & 1.7B & 4B & 0.6B & 1.7B & 4B & 0.6B & 1.7B & 4B & 0.6B & 1.7B & 4B & 0.6B & 1.7B & 4B & 0.6B & 1.7B & 4B \\
\midrule
\multicolumn{19}{l}{\cellcolor{topologyheader}\emph{Sequential}} \\
Receiver-Only & 0.622 & 0.531 & 0.518 & 0.541 & 0.587 & 0.678 & 0.570 & 0.588 & 0.720 & 0.495 & 0.516 & 0.531 & 0.508 & 0.486 & 0.686 & 0.512 & 0.486 & 0.688 \\
Receiver-Text & 0.668 & 0.620 & \underline{0.674} & 0.622 & \underline{0.603} & 0.710 & 0.648 & \textbf{0.622} & \underline{0.731} & \textbf{0.526} & 0.568 & 0.610 & \textbf{0.536} & 0.565 & 0.705 & \textbf{0.524} & 0.561 & 0.706 \\
Free Learner & \underline{0.687} & \underline{0.682} & 0.671 & \underline{0.633} & 0.584 & 0.688 & \underline{0.657} & 0.560 & 0.691 & 0.498 & \underline{0.597} & 0.657 & 0.511 & \underline{0.572} & 0.696 & 0.501 & \underline{0.571} & 0.704 \\
PropUQ-MAS & 0.620 & 0.641 & 0.554 & 0.382 & 0.494 & \textbf{0.756} & 0.395 & 0.514 & \textbf{0.783} & 0.500 & 0.548 & \underline{0.680} & 0.473 & 0.482 & \textbf{0.737} & 0.471 & 0.495 & \textbf{0.750} \\
\cmidrule(lr){1-19}
IE-Prop (Ours) & \textbf{0.711} & \textbf{0.708} & \textbf{0.684} & \textbf{0.660} & \textbf{0.622} & \underline{0.716} & \textbf{0.671} & \underline{0.610} & 0.723 & \underline{0.521} & \textbf{0.613} & \textbf{0.690} & \underline{0.533} & \textbf{0.584} & \underline{0.728} & \underline{0.518} & \textbf{0.584} & \underline{0.743} \\
\midrule
\multicolumn{19}{l}{\cellcolor{topologyheader}\emph{Fully connected}} \\
Receiver-Only & 0.616 & 0.593 & 0.507 & 0.591 & 0.477 & 0.534 & 0.603 & 0.488 & 0.457 & 0.493 & 0.509 & 0.591 & \underline{0.550} & 0.457 & 0.526 & \underline{0.562} & 0.459 & 0.557 \\
Receiver-Text & 0.696 & 0.716 & 0.652 & 0.651 & 0.682 & 0.665 & 0.655 & 0.670 & 0.676 & 0.488 & \underline{0.571} & 0.594 & 0.529 & 0.560 & 0.552 & 0.534 & 0.554 & 0.574 \\
Free Learner & \underline{0.718} & \underline{0.780} & \underline{0.719} & \underline{0.704} & \underline{0.745} & 0.765 & \underline{0.723} & \underline{0.760} & 0.782 & \underline{0.512} & 0.557 & 0.665 & 0.485 & \underline{0.577} & 0.669 & 0.495 & \underline{0.583} & 0.707 \\
PropUQ-MAS & 0.667 & 0.754 & 0.613 & 0.536 & 0.636 & \underline{0.784} & 0.547 & 0.677 & \textbf{0.819} & 0.500 & 0.547 & \textbf{0.699} & \textbf{0.567} & 0.544 & \textbf{0.695} & \textbf{0.599} & 0.564 & \textbf{0.742} \\
\cmidrule(lr){1-19}
IE-Prop (Ours) & \textbf{0.739} & \textbf{0.799} & \textbf{0.733} & \textbf{0.727} & \textbf{0.763} & \textbf{0.799} & \textbf{0.743} & \textbf{0.779} & \underline{0.812} & \textbf{0.529} & \textbf{0.573} & \underline{0.692} & 0.509 & \textbf{0.595} & \underline{0.690} & 0.519 & \textbf{0.602} & \underline{0.727} \\
\midrule
\multicolumn{19}{l}{\cellcolor{topologyheader}\emph{Hierarchy}} \\
Receiver-Only & 0.560 & 0.712 & 0.564 & 0.636 & 0.568 & 0.749 & 0.652 & 0.527 & 0.783 & 0.498 & 0.591 & 0.672 & \underline{0.510} & 0.599 & 0.682 & \underline{0.511} & 0.600 & 0.725 \\
Receiver-Text & 0.709 & 0.776 & 0.762 & 0.731 & 0.664 & 0.774 & 0.740 & 0.659 & 0.785 & \textbf{0.502} & \underline{0.628} & 0.694 & 0.478 & 0.609 & 0.683 & 0.491 & 0.615 & 0.715 \\
Free Learner & \underline{0.723} & \underline{0.780} & \underline{0.771} & \underline{0.736} & \underline{0.779} & 0.855 & \underline{0.752} & \underline{0.815} & 0.872 & 0.480 & 0.626 & 0.700 & 0.471 & \underline{0.682} & \underline{0.752} & 0.501 & \underline{0.677} & 0.769 \\
PropUQ-MAS & 0.671 & 0.710 & 0.563 & 0.639 & 0.725 & \underline{0.865} & 0.653 & 0.748 & \textbf{0.891} & 0.471 & 0.574 & \underline{0.704} & \textbf{0.514} & 0.587 & 0.750 & 0.484 & 0.575 & \underline{0.780} \\
\cmidrule(lr){1-19}
IE-Prop (Ours) & \textbf{0.748} & \textbf{0.802} & \textbf{0.784} & \textbf{0.759} & \textbf{0.794} & \textbf{0.871} & \textbf{0.769} & \textbf{0.835} & \underline{0.886} & \underline{\textbf{0.501}} & \textbf{0.645} & \textbf{0.720} & 0.490 & \textbf{0.700} & \textbf{0.772} & \textbf{0.523} & \textbf{0.696} & \textbf{0.788} \\
\bottomrule
\end{tabular}}
\caption{Final-node AUROC on GSM8K and MedQA by base-model size (Qwen3 0.6B, 1.7B and 4B): in-domain settings, seed mean per model size. Columns are grouped by dataset, then uncertainty signal, then model size. \emph{Bold:} best value in each column within each topology. \emph{Underline:} second best value in each column within each topology.}
\label{tab:main}
\end{table*}

\textbf{IE-Prop.} Our model from Eq.~\ref{eq:np}, which uses \(\hat r_p\) to mix the interaction effects for correct and incorrect sender states.

\textbf{Recursion Control.} To isolate the contribution of recursively propagated risk, we train variants of IE-Prop in which the propagated parent risk \(\hat r_p\) is replaced by the parent's local risk \(\sigma(b_p)\). This comparison isolates whether recursively accumulated upstream information improves downstream risk estimation beyond the immediate parent's local evidence and observed sender–receiver interaction.

As an external reference we also report PropUQ-MAS, computed from each agent's verbalized uncertainty, MSP, and entropy with our implementation of its propagation rule. We report AUROC for the final node. The description of the considered local risk estimates and metric is provided in Appendix ~\ref{app:local_uncertainty} and Appendix~\ref{app:eval_metric} respectively.

These comparisons address whether upstream interaction information adds predictive signal beyond what is available from the receiver's text alone, and how our interaction-conditioned approach performs against existing propagation-based uncertainty methods.

\section{Experimental Setup}
\label{sec:setup}

\paragraph{Systems.}
We evaluate three distinct four-agent topologies instantiated with Qwen3 models \citep{yang2025qwen3} (0.6B, 1.7B, and 4B variants): a \emph{sequential} pipeline where each agent observes only its immediate predecessor; a \emph{fully connected} architecture where agents attend to all preceding outputs; and a \emph{specialist hierarchy} where a final summarizer aggregates independent parallel responses. Full architectural details, including specific agent roles and prompt-level mitigations for degenerate copying behavior in the fully connected system, are deferred to Appendix~\ref{app:qa}.

\paragraph{Tasks and labels.}
We evaluate across three question-answering benchmarks: GSM8K \citep{cobbe2021gsm8k}, MedQA \citep{jin2020medqa}, and the text-only subset of WebQA \citep{chang2021webqa}. Treating each dataset, graph topology, and model-size combination as an independent setting yields 27 distinct experimental evaluations for each methods. To train and evaluate the risk estimators, each agent is assigned a binary correctness label by extracting the final answer from its visible output and comparing it against the ground truth. Full details regarding train-test data splits, automated answer extraction protocols (e.g., deterministic vs. LLM-based read-outs), and task-specific matching criteria are provided in Appendix~\ref{app:qa}.

\paragraph{In-Domain.}
Estimators are trained on the execution traces from a dataset's training split and evaluated on its corresponding held-out test split. Detailed in-domain results are provided in \S\ref{sec:results}.

\paragraph{Transfer.}
To evaluate whether the learned risk signals generalize to unseen distributions, we employ a leave-one-dataset-out transfer protocol. For each architecture and model size, estimators are trained on execution traces from two datasets and evaluated on the held-out third dataset. Detailed cross-dataset transfer results are in Appendix~\ref{app:res-transfer}.

\section{Results}
\label{sec:results}

We evaluate whether realized inter-agent exchanges improve downstream error-risk estimation, whether sender-state conditioning contributes beyond an unconstrained interaction model. We then examine downstream error discrimination with correct upstream context and isolate the contribution of recursive risk propagation. \tabref{tab:main} reports seed-averaged final-node AUROC for Qwen3-0.6B, 1.7B, and 4B across datasets.

\subsection{Interaction information beyond receiver text}
\label{sec:res-chain}

The receiver's visible output provides useful evidence for error-risk estimation, but does not exhaust the predictive information available in the execution trace. In \tabref{tab:main}, IE-Prop exceeds Receiver-Text in 46 of the 54 topology--dataset--uncertainty--model-size combinations. Averaged over the datasets and uncertainty signals, the gains for the 0.6B, 1.7B, and 4B models are respectively +0.015, +0.030, and +0.025 in sequential systems; +0.036, +0.060, and +0.123 in fully connected systems; and +0.023, +0.087, and +0.068 in the hierarchy. Interaction information therefore improves mean discrimination at each evaluated size within all three topologies, although magnitude of the gain depends on setting. 

\begin{figure}[!t]
    \centering
    \includegraphics[
        width=\columnwidth,
        height=\textheight,
        keepaspectratio
    ]{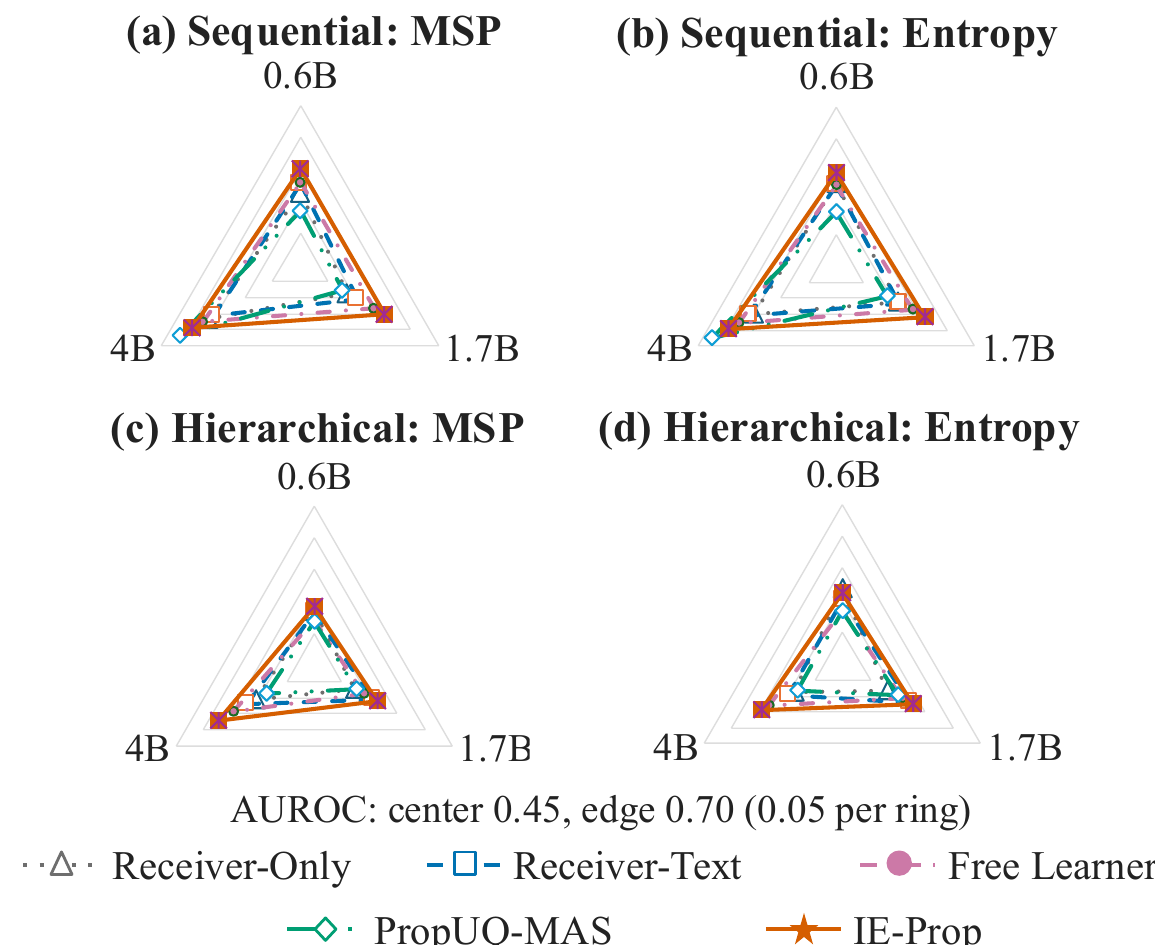}
    \caption{Final-node AUROC on WebQA for sequential and hierarchical systems using MSP and entropy across Qwen3 model sizes, averaged over seeds.}
    \label{fig:webqa}
\end{figure}

\figref{fig:webqa} shows that IE-Prop also improves over Receiver-Text on WebQA in all 12 evaluated settings, achieving the highest AUROC in nine. It leads in most hierarchical settings and at the two smaller model sizes in sequential systems. However, PropUQ-MAS performs best at 4B in sequential systems for both MSP and entropy, indicating that the relative advantage depends on architecture and model size.

\subsection{State-conditioned effects}
\label{sec:res-recursion}

We next compare IE with the Free Learner to examine whether the structured sender-state-conditioned update improves on an unconstrained interaction adjustment. In \tabref{tab:main}, IE achieves higher seed-averaged AUROC in all 54 displayed combinations, covering every model size, topology, dataset, and uncertainty signal. Averaged over datasets, uncertainty signals, and model sizes, the gains are approximately +0.026 in the sequential system, +0.021 in the fully connected system, and +0.019 in the hierarchy.

The consistency of this comparison across model sizes supports the state-conditioned formulation as a useful inductive bias for the evaluated GSM8K and MedQA settings. This predictive advantage does not, by itself, establish that the two learned heads identify distinct error-propagation and correction mechanisms; it shows that the structured formulation improves error-risk discrimination relative to the Free Learner. Additional experiments with oracle-supervised sender-state conditioning are presented in Appendix~\ref{app:oracle_sender_state}.

\begin{figure}[!t]
    \centering
    \includegraphics[
        width=\columnwidth,
        height=\textheight,
        keepaspectratio
    ]{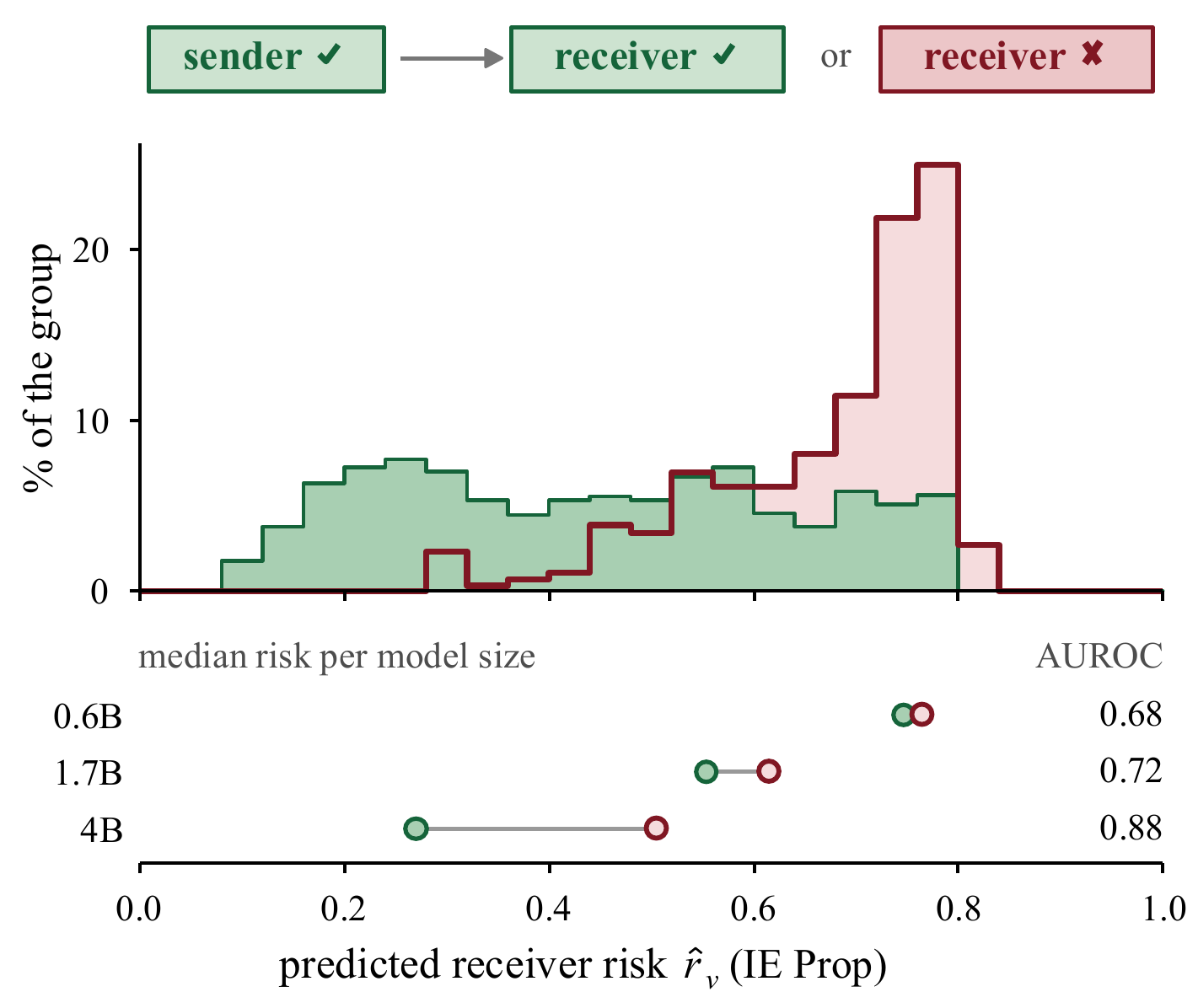}
    \caption{Predicted receiver risk on sequential MedQA traces with correct upstream answers. \emph{Top:} Risk distributions by receiver correctness. \emph{Bottom:} Median risk and AUROC across Qwen3 model sizes.}
    \label{fig:reruns}
\end{figure}

\subsection{Downstream risk discrimination with correct upstream context}
\label{sec:risk-discrimination}

A correct upstream response does not guarantee that the receiver will remain correct. \figref{fig:reruns} examines sequential MedQA interactions restricted to cases where the sender's answer is correct. Within this subset, IE assigns higher predicted risk to incorrect receivers than to correct receivers at each evaluated model size.

Discrimination improves across the three sizes in this diagnostic, with AUROC values of 0.68, 0.72, and 0.88 for the 0.6B, 1.7B, and 4B models, respectively. These results show that the estimator can distinguish downstream errors even when the upstream answer is correct. They do not establish independence from estimated sender risk, since the model still uses that estimate when combining its interaction effects. Rather, they demonstrate that its predictive utility extends to interactions in which downstream failure cannot be attributed simply to an incorrect upstream answer.

\subsection{Contribution of recursive risk propagation}
\label{sec:recursion}

\begin{figure}[!t]
    \centering
    \includegraphics[
        width=\columnwidth,
        height=\textheight,
        keepaspectratio
    ]{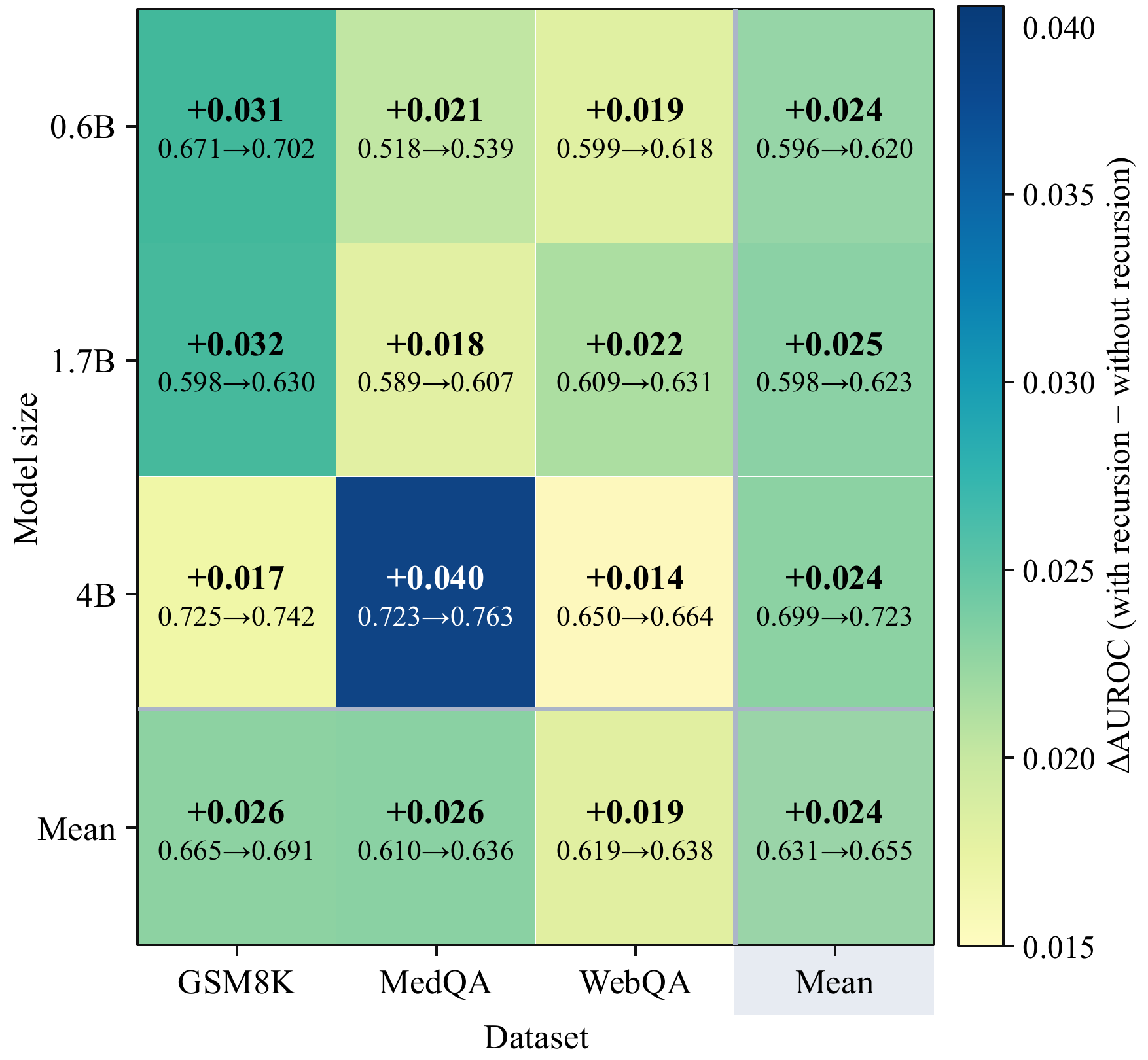}
    \caption{Final-node AUROC gains from recursive risk propagation in the sequential system using Qwen3 models. Cells show gains and corresponding scores (without → with recursion), averaged over three seeds.}
    \label{fig:recursive}
\end{figure}

Finally, we isolate the contribution of recursive propagation by comparing IE with a matched variant that replaces recursively estimated parent risk with local parent risk. In the sequential ablation shown in \figref{fig:recursive}, recursion improves final-node AUROC in all nine displayed dataset--model-size combinations. The overall mean increases from 0.631 to 0.655 (+0.024).

Averaged over datasets, the gains are +0.024, +0.025, and +0.024 for the 0.6B, 1.7B, and 4B models, respectively. Thus, the average benefit is similar across model sizes, although individual improvements vary. The largest gain occurs on MedQA with the 4B model (+0.040), while gains averaged over model sizes are +0.026 on GSM8K, +0.026 on MedQA, and +0.019 on WebQA. These results indicate that recursively accumulated parent risk provides predictive information beyond local parent uncertainty and immediate sender--receiver interaction in the evaluated sequential systems. 

\section{Discussion}
\label{sec:discussion}

Our results suggest that downstream error risk depends not only on upstream correctness but also on how receivers respond to upstream information. As shown in \secref{sec:risk-discrimination}, IE-Prop distinguishes downstream errors even when upstream answers are correct, illustrating why sender correctness alone is insufficient to characterize downstream risk. Moreover, its improvement over the Free Learner (\secref{sec:res-recursion}) supports sender-state-conditioned interaction modeling as a useful predictive inductive bias. The benefit of recursive propagation depends on the system's information-flow structure. In sequential systems, earlier interactions reach the final agent only through intermediate agents, making recursively accumulated risk informative beyond the immediate interaction (\secref{sec:recursion}). In contrast, in shallow hierarchies, the final agent receives information directly from independent specialists, allowing interaction modeling to improve risk discrimination without multi-hop recursive accumulation. These findings suggest interaction modeling and recursive risk propagation provide complementary predictive benefits, whose relative importance depends on system topology.


\section{Conclusion}
\label{sec:conclusion}

In this work, we investigated post-hoc agent-level error-risk estimation using execution traces from LLM-based multi-agent systems. We introduced IE-Prop, an approach that models the contribution of realized sender–receiver interactions to downstream error risk by learning separate effects conditioned on whether the sender is correct or incorrect. By combining these effects using estimated sender risk and recursively propagating the resulting estimates through the execution graph, our method captures interaction-dependent error dynamics beyond monotone propagation assumptions. Across three multi-agent architectures and three question-answering benchmarks, our approach generally improves downstream error-risk discrimination over receiver-local and existing propagation-based baselines. Our analyses further demonstrate that interaction information remains predictive even when upstream answers are correct, while recursive propagation provides additional benefits in sequential systems. Together, these findings underscore the value of modeling realized inter-agent interactions for downstream error-risk estimation.

\section*{Limitations}

Our evaluation focuses on three four-agent architectures, three question-answering benchmarks, and one model family, leaving generalization to larger and more diverse multi-agent systems for future work. Differences in agent roles and prompting protocols also limit direct comparisons across architectures. Although interaction modeling generally improves error-risk discrimination, the gains vary across settings, and the distinct contributions of sender-state conditioning and recursive propagation are not fully established in every topology. Furthermore, our estimator operates post-hoc and relies partly on automated correctness labels. Finally, the additive interaction model approximates marginal inference, and its learned effects represent predictive associations rather than causal relationships. Future work could explore broader agentic settings, alternative aggregation mechanisms, and more direct supervision of sender-state effects.

\section*{Ethical Considerations}

This work investigates error-risk estimation in LLM-based multi-agent systems, with the goal of improving their reliability. However, inaccurate risk estimates may create a false sense of confidence in incorrect outputs, particularly in high-stakes domains such as medical question answering. Our method is evaluated on established benchmarks and does not guarantee correctness or safety in real-world deployments. Additionally, the learned risk estimates may not generalize reliably to unseen domains, model families, or multi-agent architectures. Finally, the use of multiple LLM agents and additional risk-estimation models incurs computational costs and associated environmental impacts.

\section*{Computational Resources.}
All experiments were conducted using a computing infrastructure equipped with four NVIDIA A100 GPUs. These GPUs supported the complete experimental workload, including LLM hosting, multi-agent inference, execution trace generation, and training of the risk-estimation networks.


\bibliography{references}

\clearpage
\appendix

\section{Approximation Bound}
\label{app:proof}
Let $Z_v=b_v+\sum_p\delta^{(E_p)}_{pv}$, where the parent states $E_{\Pa(v)}$ follow some distribution $\pi$, and let $\mu=\mathbb{E}_\pi[Z_v]$.
The exact marginal under the conditional model of Eq.~\ref{eq:cpd} is $\mathbb{E}_\pi[\sigma(Z_v)]$, and Eq.~\ref{eq:np} computes $\sigma(\mu)$ when given the exact parent marginals.

\paragraph{Bound.}
Taylor's theorem with the Lagrange remainder gives $\sigma(Z)=\sigma(\mu)+\sigma'(\mu)(Z-\mu)+\tfrac12\sigma''(\xi)(Z-\mu)^2$ for some $\xi$ between $\mu$ and $Z$.
Taking expectations, the linear term vanishes, so $|\mathbb{E}[\sigma(Z)]-\sigma(\mu)|\le\tfrac12\sup_z|\sigma''(z)|\,\mathrm{Var}(Z)$.
With $\sigma''=\sigma(1-\sigma)(1-2\sigma)$, the supremum is attained at $\sigma=\tfrac12\pm\tfrac{1}{2\sqrt3}$ and equals $1/(6\sqrt3)$, so
\begin{equation}
  \big|\mathbb{E}[\sigma(Z_v)]-\sigma(\mathbb{E}[Z_v])\big|\ \le\ \frac{1}{12\sqrt3}\,\mathrm{Var}(Z_v).
  \label{eq:bound}
\end{equation}

\paragraph{Assumptions and interpretation.}
The bound uses only the first two moments of $Z_v$, so it holds for any parent-state distribution, including correlated ones.
Writing $\Delta_{pv}=\done-\dzero$, we have $Z_v=b_v+\sum_p\dzero+\sum_pE_p\Delta_{pv}$ and
\begin{multline*}
  \mathrm{Var}(Z_v)=\sum_p\Delta_{pv}^2\,r_p(1-r_p)\\
  +2\sum_{p<q}\Delta_{pv}\Delta_{qv}\,\mathrm{Cov}(E_p,E_q).
\end{multline*}
The error therefore vanishes when the two effects coincide, and grows with strong, uncertain and positively co-varying parent effects; correlations do not enter the update itself but change the variance that controls its error.
Exactness holds whenever $\mathrm{Var}(Z_v)=0$: when the parent states are known, when $\dzero=\done$ on every uncertain edge, or when every parent configuration with positive probability gives the same logit.
The bound is local: it compares the update and exact marginalization under the same parent distribution and does not account for errors in propagated parent marginals.

\paragraph{Single parent.}
For $k=1$, as in our sequential system, $\mathrm{Var}(Z_v)=\Delta_{pv}^2\,r_p(1-r_p)\le\Delta_{pv}^2/4$, so the error is at most $\Delta_{pv}^2/(48\sqrt3)$ and is largest for parents of maximal uncertainty.
The final agent of the fully connected system has $k=3$ parents that share the first agent as an ancestor, so the covariance term applies there.

\section{Natural-Parameter Interpretation and Training Analysis}
\label{app:np}

\subsection{Reverse-KL barycenter}
\label{app:kl}
Each parent configuration $e$ induces a receiver distribution $p_e=\mathrm{Bernoulli}(\sigma(\eta_e))$ with $\eta_e=b_v+\sum_p\delta^{(e_p)}_{pv}$.
Consider summarizing the collection $\{p_e\}$ by the single Bernoulli $q=\mathrm{Bernoulli}(r)$ that minimizes $\mathbb{E}_{e\sim\pi}[D_{\mathrm{KL}}(q\Vert p_e)]$.
Since $\partial_r D_{\mathrm{KL}}(q\Vert p_e)=\operatorname{logit}(r)-\eta_e$ and the objective is strictly convex, the minimizer satisfies $\operatorname{logit}(r)=\mathbb{E}_\pi[\eta_e]$, i.e.\ $r=\sigma(\mathbb{E}[Z_v])$.
With the additive natural parameter, $\mathbb{E}_\pi[\eta_e]=b_v+\sum_p[(1-r_p)\dzero+r_p\done]$, which depends on $\pi$ only through the individual parent marginals $r_p$, for any dependence between parents.
This is an interpretation of the scalar composition rule, not a justification that it yields the exact risk, and not a claim that it is empirically preferable to other operators.
The exact mixture over a binary variable is itself Bernoulli with mean $\mathbb{E}_\pi[\sigma(\eta_e)]$; the update instead aggregates the state-conditioned distributions in reverse-KL (natural-parameter) geometry.
Eq.~\ref{eq:np} further replaces the exact parent marginals by propagated ones, so its error combines the local term of Eq.~\ref{eq:bound} with errors accumulated upstream.

\section{Induced Probabilistic Model}
\label{app:graphical}
Conditioned on a completed trace $X$, the node-wise conditionals of Eq.~\ref{eq:cpd} can be combined into the DAG factorization $P(E\mid X)=\prod_vP(E_v\mid E_{\Pa(v)},X)$, a Bayesian network over error indicators whose structure is the communication graph.
This is a modeling assumption about dependencies among correctness variables given the trace, not a claim about the causal structure of failures, and it does not model how messages were generated.
The deployed estimator does not maintain this joint model; it propagates only the scalar risks of Eq.~\ref{eq:np}.

\paragraph{Relation to sigmoid belief networks and log-linear models.}
Each conditional is a logistic regression on parent-state indicators, i.e.\ a sigmoid belief network \citep{neal1992connectionist} whose weights $\Delta_{pv}$ and intercept $b_v+\sum_p\dzero$ are computed from the trace.
Equivalently, the parent-state indicators enter a log-linear model of the receiver's error with first-order interaction terms only \citep{besag1974spatial}.
Noisy-OR \citep{pearl1988probabilistic}, the contamination rule, replaces the logistic link by a product of per-parent failure terms, so a wrong parent can only increase the receiver's risk.

\paragraph{Exact marginalization and parent dependence.}
The exact marginal of a receiver is $\sum_e\pi_v(e)\,\sigma(\eta_v(e))$, where $\pi_v$ is the joint distribution of its parents' states.
Parents that share ancestors are dependent under the induced network even when the conditionals are fixed, so $\pi_v$ is not a product of marginals in general.
Exact inference must sum over the joint states of the latent ancestors that a receiver depends on.
Variable elimination reduces this cost to exponential in the induced width of the elimination order, but a $k$-parent logistic receiver still requires the $2^k$-entry joint of its parents, and overlapping ancestry forces joint tables over shared ancestors.
In the controlled sequential system every receiver has one parent, and in the fully connected system the final agent's three dependent parents have only eight joint states, so exact marginalization is cheap in both; the update matters for wider graphs.


\section{Controlled QA: Details and Full Results}
\label{app:qa}\label{app:persetting}\label{app:prrsteps}
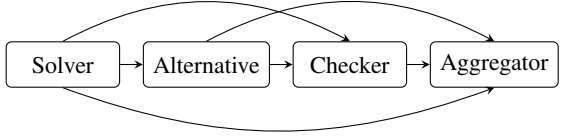
\begin{figure}[t]
\centering
\begin{tikzpicture}[>=stealth, node distance=3mm, ag/.style={draw, rounded corners=2pt, minimum width=15mm, minimum height=6mm, font=\footnotesize, align=center}]
\node[ag] (a1) {Solver};
\node[ag, right=of a1] (a2) {Alternative};
\node[ag, right=of a2] (a3) {Checker};
\node[ag, right=of a3] (a4) {Aggregator};
\draw[->] (a1) -- (a2);
\draw[->] (a2) -- (a3);
\draw[->] (a3) -- (a4);
\draw[->] (a1.north) to[bend left=28] (a3.north);
\draw[->] (a2.north) to[bend left=28] (a4.north);
\draw[->] (a1.south) to[bend right=20] (a4.south);
\end{tikzpicture}
\caption{The fully connected system. Every agent reads the visible outputs of all earlier agents (six edges). The Alternative-Method agent and the Checker first write a private draft without the earlier outputs; the draft is never passed on.}
\label{fig:fc}
\end{figure}

\FloatBarrier
\paragraph{Labels.}
Each agent's label is derived from the answer stated in its visible output, after hidden reasoning and uncertainty tags are removed, and the same procedure labels training and test traces; training uses every node's label as a target, not only the final node's.
On GSM8K the answer is the last boxed number, or otherwise the last number in the text, and it is compared with the gold number.
On MedQA and WebQA the answer is extracted by a read-out with the same Qwen3 model that generated the trace: given the question, the agent's role and its visible response, but not the gold answer, the model returns a JSON object naming the option letter (MedQA) or the complete-sentence answer (WebQA) that the response supports, or UNKNOWN, and it is instructed not to solve the question itself.
Decoding is greedy with thinking disabled and a short output budget, and an invalid JSON response is retried once.
Correctness is decided in code: exact option match for MedQA, and full recall of normalized gold-answer tokens against any accepted answer for WebQA.
A node whose response supports no answer is unlabeled, and training traces with any unlabeled node are dropped.
In the hierarchy, the Summarizer's label is the system evaluator's verdict; in the fully connected system every agent is labeled as in the sequential system.
Test items whose final node is unlabeled are dropped, and oracle evaluations additionally drop items with an unlabeled parent; oracle predictions were not computed for the sequential MedQA 4B setting.

Hyperparameters are shared by every cell: hidden width 256, one hidden layer, dropout 0.1, AdamW with learning rate $10^{-3}$ and weight decay $10^{-4}$, cosine schedule, batch 256, at most 30 epochs with patience 5, and five question-level folds; each fold model is selected by validation final-node AUROC and the five fold predictions are averaged at test time.
The oracle numbers should not be read as attainable targets, since parent and final-node labels largely coincide.



\section{Cross-dataset transfer}
\label{app:res-transfer}

\begin{figure*}[t]
\centering
\includegraphics[width=\textwidth]{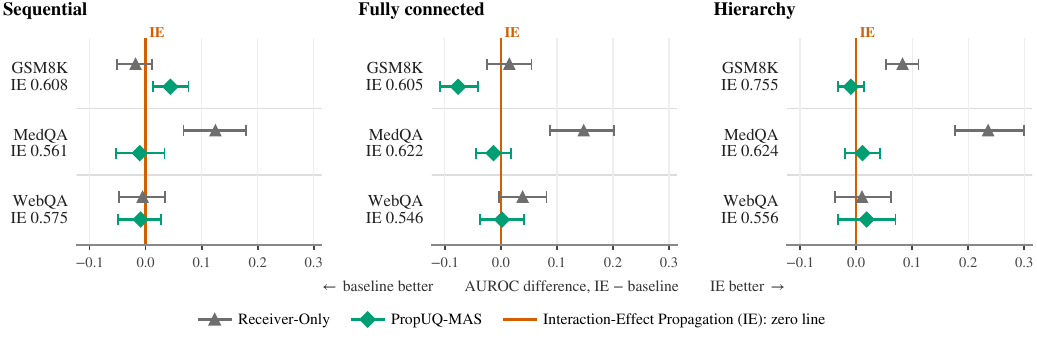}
\caption{Cross-dataset transfer performance across three multi-agent topologies, showing final-node AUROC and differences relative to Receiver-Only and PropUQ-MAS baselines.}
\label{fig:lodo-forest}
\end{figure*}

Figure~\ref{fig:lodo-forest} evaluates leave-one-dataset-out transfer with entropy-based uncertainty: learned estimators are trained on traces from two datasets and evaluated on the held-out third dataset. It reports IE's final-node AUROC and its difference from Receiver-Only and the training-free PropUQ-MAS. The held-out IE AUROCs are 0.608, 0.561, and 0.575 in the sequential system; 0.605, 0.622, and 0.546 in the fully connected system; and 0.755, 0.624, and 0.556 in the hierarchy, for GSM8K, MedQA, and WebQA, respectively.

IE improves over Receiver-Only most clearly on held-out MedQA across all three graphs. Its advantage over PropUQ-MAS is less consistent: the paired intervals overlap zero in seven of the nine graph--held-out-dataset comparisons, while sequential GSM8K favors IE and fully connected GSM8K favors PropUQ-MAS. Transfer therefore offers evidence that learned interaction information can remain useful outside its training datasets, but not that IE uniformly dominates a propagation-based uncertainty reference. Because Figure~\ref{fig:lodo-forest} does not include Receiver-Text or matched recursion controls, it does not establish that the transfer improvements arise from state-conditioned recursion or exceed a receiver-text predictor.

\section{Oracle-Supervised Sender-State Conditioning}
\label{app:oracle_sender_state}

The original interaction-effect (IE) objective supervises the sender-state heads
$\delta_{pv}^{(0)}$ and $\delta_{pv}^{(1)}$ only through their risk-weighted
mixture. Consequently, good predictions need not imply that the heads are
aligned with sender correctness. To test this, we add an oracle-conditioned
training loss that selects the head corresponding to the observed sender label:
\begin{equation}
\begin{aligned}
\hat{E}_{iv}^{\mathrm{oracle}}
  &= \sigma\left(
     b_{iv} +
     \sum_{p\in\mathrm{Pa}(v)}
     \delta_{ipv}^{(E_{ip})}
     \right),\\
\mathcal{L}_{\mathrm{oracle}}
  &= \frac{1}{4N}
     \sum_{i=1}^{N}
     \sum_{v\in\mathcal{V}_i^{\mathrm{eligible}}}
     \operatorname{BCE}
     \left(\hat{E}_{iv}^{\mathrm{oracle}}, E_{iv}\right),\\
\mathcal{L}_{\mathrm{total}}
  &= \mathcal{L}_{\mathrm{pred}}
     + \lambda\mathcal{L}_{\mathrm{oracle}}.
\end{aligned}
\end{equation}
where $\mathcal{V}_{i}^{\mathrm{eligible}}$ contains non-root nodes with known
receiver and parent labels. The original $\mathcal{L}_{\mathrm{pred}}$ is
unchanged and uses the same $1/(4N)$ normalization. At test time, inference
uses the original soft recursive mixture without sender labels.

We evaluate all three architectures on GSM8K, MedQA, and WebQA, across three
Qwen3 sizes and three uncertainty signals (three seeds; five question-level
folds per seed). We select $\lambda=0.1$ from
$\{0,0.1,0.3,1,3\}$ by mean out-of-fold final-node AUROC on training data.
Relative to the original IE model, the mean change in deployed AUROC is
$+0.004$ (sequential), approximately $0$ (fully connected), and $-0.001$
(hierarchy). Thus, oracle supervision leaves predictive performance largely
unchanged.

To assess head alignment, we compute the oracle-state swap gap:
$G_{\mathrm{swap}}=\operatorname{AUROC}(E_p)-
\operatorname{AUROC}(1-E_p)$, where each receiver uses either the observed or
flipped sender labels to select its incoming heads. With $\lambda=0$, the
mean gaps are $+0.271$, $-0.050$, and $-0.283$ for sequential, fully
connected, and hierarchical systems, respectively; at $\lambda=0.1$, they
increase to $+0.570$, $+0.740$, and $+0.679$. These oracle diagnostics use
cross-fitted training traces that also served in fold-level validation;
independent test traces corroborate the trend in the sequential system.
Swapping estimated sender risks in the deployed soft rule likewise produces
a larger AUROC gap after oracle supervision. These results support improved
\emph{sender-state alignment}, but do not identify causal error-transmission
or correction effects: strong sender--receiver correctness correlations can
also drive large swap gaps.

\section{Local Uncertainty Signals}
\label{app:local_uncertainty}

We consider three agent-local uncertainty signals:
verbalized uncertainty, Maximum Sequence Probability
(MSP), and prediction entropy. Each signal is evaluated
independently as an input to the local risk estimator
$g_\phi$ in Eq.~\ref{eq:local}. Let $h_v$ denote
the generation context of agent $v$, and let
$y_v=(y_{v,1},\ldots,y_{v,T_v})$ denote the generated
token sequence used for uncertainty estimation.

\paragraph{Verbalized uncertainty.}
Verbalized uncertainty is obtained from the model's
self-reported confidence in its generated answer
\citep{lin2022words,tian2023calibration,xiong2024express}.
Let $c_v\in[0,1]$ denote the confidence expressed by
agent $v$, where larger values indicate greater
confidence. We convert this to an uncertainty score as
\begin{equation}
    u_v^{\mathrm{verb}} = 1-c_v.
\end{equation}
Thus, larger values indicate greater self-reported
uncertainty.

\paragraph{Maximum Sequence Probability (MSP).}
MSP uses the probability assigned by the language model
to the generated token sequence. Let
$p_\mathrm{LM}(y_{v,t}\mid h_v,y_{v,<t})$ denote
the conditional probability of the generated token
$y_{v,t}$. We compute the length-normalized sequence
probability as
\begin{equation}
    s_v^{\mathrm{MSP}}
    = \exp\left(
    \frac{1}{T_v}\sum_{t=1}^{T_v}
    \log p_\mathrm{LM}(y_{v,t}\mid h_v,y_{v,<t})
    \right).
\end{equation}
The corresponding uncertainty score is
\begin{equation}
    u_v^{\mathrm{MSP}} = 1-s_v^{\mathrm{MSP}}.
\end{equation}
Length normalization reduces the dependence of the
score on response length, while lower sequence
probability indicates greater uncertainty.

\paragraph{Prediction entropy.}
Prediction entropy measures uncertainty in the model's
next-token distributions. Let
$p_{v,t}(k)$ denote the predicted probability of token
$k$ at generation step $t$, and let $\mathcal{V}$
be the vocabulary. The token-level predictive entropy is
\begin{equation}
    H_{v,t} = -
    \sum_{k\in\mathcal{V}}
    p_{v,t}(k)\log p_{v,t}(k).
\end{equation}
We aggregate token-level entropies over the generated
sequence:
\begin{equation}
    u_v^{\mathrm{ent}}
    = \frac{1}{T_v}\sum_{t=1}^{T_v}H_{v,t}.
\end{equation}
Higher entropy indicates a more dispersed predictive
distribution and therefore greater uncertainty.
Unlike semantic entropy
\citep{kuhn2023semantic,farquhar2024detecting},
this signal operates directly on token-level
distributions without grouping sampled responses
by semantic equivalence.

\paragraph{Integration with local risk estimation.}
For each uncertainty signal
$s\in\{\mathrm{verb},\mathrm{MSP},\mathrm{ent}\}$,
we use the corresponding scalar $u_v^{(s)}$
as the agent-local input $w_v$ to the learned
local estimator:
\begin{equation}
    b_v = g_\phi(u_v^{(s)}).
\end{equation}
The local error-risk estimate is $\sigma(b_v)$,
which serves as the Receiver-Only prediction and
provides the local component of the interaction-effect
propagation model. All three signals are evaluated
separately under the same experimental protocol.

\section{Evaluation Metric}
\label{app:eval_metric}

We evaluate error-risk estimation using the Area Under the
Receiver Operating Characteristic Curve (AUROC). For each
evaluated final-node response $i$, let $E_i \in \{0,1\}$
denote its correctness label, where $E_i=1$ indicates an
incorrect answer and $E_i=0$ indicates a correct answer.
Let $\hat{r}_i \in [0,1]$ denote the corresponding predicted
error risk.

The receiver operating characteristic (ROC) curve plots
the true positive rate (TPR) against the false positive
rate (FPR) as the decision threshold $\tau$ varies:
\begin{equation}
\begin{aligned}
    \mathrm{TPR}(\tau)
    &= P(\hat{r}_i \geq \tau \mid E_i=1), \\
    \mathrm{FPR}(\tau)
    &= P(\hat{r}_i \geq \tau \mid E_i=0).
\end{aligned}
\end{equation}
AUROC summarizes the area under this curve and can
equivalently be interpreted as
\begin{equation}
    \mathrm{AUROC}
    = P(\hat{r}^{+} > \hat{r}^{-})
    + \frac{1}{2}P(\hat{r}^{+} = \hat{r}^{-}),
\end{equation}
where $\hat{r}^{+}$ and $\hat{r}^{-}$ are the predicted
risks of randomly selected incorrect and correct
responses, respectively.

An AUROC of $0.5$ corresponds to random discrimination,
while $1.0$ indicates perfect separation between
incorrect and correct responses. Higher AUROC therefore
indicates better discrimination of downstream errors.
Unlike threshold-dependent classification metrics,
AUROC evaluates the ranking of predicted risks across
all decision thresholds, without requiring a fixed
operating threshold. It measures discrimination rather
than probability calibration. We compute AUROC on
held-out final-node responses and report seed-averaged
results where applicable.

\begin{algorithm}[t]
\small
\caption{Interaction-effect propagation: inference, training and $K$-fold prediction.}
\label{alg:ie}
\begin{algorithmic}[1]
\Statex \textbf{Trace} $G=(V,\mathcal{E})$ for task $q$; agent $v$ has role $\rho_v$, visible output $m_v$, local signal $w_v$
\Procedure{Propagate}{$G;\,\theta,\phi,T{=}1$}
  \For{$(p,v)\in\mathcal{E}$}
    \State $x_{pv}\gets\operatorname{Enc}(q,\rho_p,m_p,\rho_v,m_v)$ \Comment{Eq.~\ref{eq:edge}; frozen}
    \State $(\delta^{(0)}_{pv},\delta^{(1)}_{pv})\gets f_\theta(x_{pv})$
  \EndFor
  \For{$v\in V$ in topological order}
    \State $b_v\gets g_\phi(w_v)$ \Comment{Eq.~\ref{eq:local}}
    \For{$p\in\mathrm{Pa}(v)$}
      \State $\bar\delta_{pv}\gets(1-\hat r_p)\,\delta^{(0)}_{pv}+\hat r_p\,\delta^{(1)}_{pv}$ \Comment{Eq.~\ref{eq:mix}}
    \EndFor
    \State $\hat r_v\gets\sigma\bigl(b_v+\sum_{p\in\mathrm{Pa}(v)}\bar\delta_{pv}\bigr)$ \Comment{Eq.~\ref{eq:np}}
  \EndFor
  \State \Return $\{\hat r_v\}_{v\in V}$ \Comment{roots: $\hat r_v=\sigma(b_v)$}
\EndProcedure
\Statex
\Procedure{Train}{$\mathcal{D}$} \Comment{labeled traces; \S\ref{sec:learning}}
  \State split $\mathcal{D}$ into $K$ question-level folds $\mathcal{D}_1,\dots,\mathcal{D}_K$
  \For{$k=1,\dots,K$}
    \State initialize $\theta_k,\phi_k$
    \Repeat \Comment{epochs; minibatches $B\subset\mathcal{D}\setminus\mathcal{D}_k$}
      \State $\{\hat r_v\}\gets\Call{Propagate}{G;\,\theta_k,\phi_k}$, $G\in B$
      \State $\mathcal{L}\gets\sum_{G\in B}\sum_{v\in V}\operatorname{BCE}(\hat r_v,E_v)$ \Comment{Eq.~\ref{eq:loss}}
      \State AdamW step on $\nabla_{\theta_k,\phi_k}\mathcal{L}$ \Comment{through all $\hat r_p$}
    \Until{final-node AUROC on $\mathcal{D}_k$ stops improving}
    \State keep the best epoch; record its node logits on $\mathcal{D}_k$
  \EndFor
  \State \Return $\{(\theta_k,\phi_k)\}_{k=1}^{K}$
\EndProcedure
\Statex
\Function{Predict}{$G$}
  \State \Return $\frac{1}{K}\sum_{k=1}^{K}\Call{Propagate}{G;\,\theta_k,\phi_k,T}$
\EndFunction
\end{algorithmic}
\end{algorithm}

\end{document}